\documentclass[runningheads]{llncs}
\usepackage[utf8]{inputenc}
\usepackage{graphicx}
\usepackage[table,xcdraw]{xcolor}
\usepackage{url}
\usepackage{float}
\usepackage{amsmath}
\usepackage{multirow}
\usepackage[caption=false]{subfig}
\usepackage{comment}
\usepackage{multicol}
\usepackage{cite}

\newcolumntype{L}[1]{>{\raggedright\let\newline\\\arraybackslash\hspace{0pt}}m{#1}}
\newcolumntype{C}[1]{>{\centering\let\newline\\\arraybackslash\hspace{0pt}}m{#1}}
\newcolumntype{R}[1]{>{\raggedleft\let\newline\\\arraybackslash\hspace{0pt}}m{#1}}
\begin{document}
\title{Efficient Lines Detection for Robot Soccer\thanks{Supported by Centro de Informática (CIn - UFPE), Fundação de Amparo a Ciência e Tecnologia do Estado de Pernambuco (FACEPE), and RobôCIn Robotics Team.}}
%
\titlerunning{Efficient Lines Detection}
%
\author{
Jo\~ao G. Melo \and
Jo\~ao P. Mafaldo \and
Edna Barros}
\authorrunning{J. Melo et al.}
%
\institute{Centro de Informática, Universidade Federal de Pernambuco. \\
Av. Prof. Moraes Rego, 1235 - Cidade Universitária, Recife - Pernambuco, Brazil.
\email{\{jgocm, jpmp, ensb\}@cin.ufpe.br}}

\authorrunning{J. Melo et al.}
%
\maketitle              

\begin{abstract}
Self-localization is essential in robot soccer, where accurate detection of visual field features, such as lines and boundaries, is critical for reliable pose estimation. This paper presents a lightweight and efficient method for detecting soccer field lines using the ELSED algorithm, extended with a classification step that analyzes RGB color transitions to identify lines belonging to the field. We introduce a pipeline based on Particle Swarm Optimization (PSO) for threshold calibration to optimize detection performance, requiring only a small number of annotated samples. Our approach achieves accuracy comparable to a state-of-the-art deep learning model while offering higher processing speed, making it well-suited for real-time applications on low-power robotic platforms.

\end{abstract}

\section{Introduction} \label{intro}


In robot soccer, computer vision plays a crucial role in detecting visual field references that assist in self-localization. The main features used include field lines, field boundaries, goalposts, and line intersections \cite{TORSO2021}. Reliable identification of these elements is essential for ensuring precise localization, as they are the primary fixed landmarks available in the environment.

Robotic systems have limited processing capabilities, and the dynamism in the robot soccer environment requires real-time responses. However, the robot must perform multiple computationally intensive tasks, such as vision processing, localization, and trajectory planning. Therefore, it is necessary to find trade-offs between computational cost and accuracy to ensure that robots can operate efficiently and competitively \cite{towards-autonomous-ssl, humanoid:vision-pipeline}.

Several authors have proposed efficient methods for detecting field features. Some approaches include segmentation based on vertical and horizontal scan lines \cite{sfrl-1}, convolutional neural networks (CNNs) for detecting field boundaries \cite{DeepFieldBoundary}, and unified models capable of performing object detection and segmentation simultaneously \cite{yoeo}. These techniques aim to improve self-localization reliability without compromising computational efficiency.

In this paper, we propose a novel method for detecting soccer field lines based on the Enhanced Line SEgment Drawing (ELSED) algorithm \cite{elsed}, which is an efficient method for detecting line segments in images. Our approach introduces an additional step in the algorithm to classify the detected segments, allowing the identification of lines belonging to the soccer field.

Additionally, our method\footnote{https://github.com/jgocm/ELSED-SSL} features adjustable thresholds, optimizing detection performance according to environmental conditions. To facilitate this adjustment, we propose a pipeline based on Particle Swarm Optimization (PSO), which enables efficient calibration of thresholds using a small number of images and annotations. With this approach, we achieve reliable results with reduced computational costs, making our solution viable for real-time applications in robot soccer. We summarize the contributions of this work as follows:
\begin{itemize}
    \item A novel and efficient approach to detect soccer field lines.
    \item A pipeline for easily collecting data, labeling line segments, and training the algorithm's classification thresholds.
    \item A performance evaluation of our method and a state-of-the-art method running in a low-power device regarding accuracy and processing speed.
\end{itemize}

\section{Related Work}\label{ch2}
In robot soccer, a traditional solution for detecting field elements relies on performing a pixel-wise segmentation on grid lines of the image and computing their gradients to classify them among the possible field features \cite{sfrl-1}. Although this method has been enhanced throughout the years, it still compounds the basis for acquiring observations for localization in the Standard Platform League (SPL) environment \cite{BHumanCodeRelease2019, Winner-RoboCup-2022}.

Other approaches propose applying color segmentation to the whole image using the HSV color space and classifying the pixels based on thresholds adjusted by humans before the competition, matching the white of the lines and goal \cite{humanoid:vision-system, humanoid:vision-pipeline}. Furthermore, Farazi et al. \cite{humanoid:vision-system} propose applying a probabilistic Hough Transform to the segmented image to extract line segments.

The previously mentioned methods require adjusting thresholds before matches. To overcome this limitation, SPL researchers propose a lighting-independent field line detection, which calculates all needed parameters online \cite{Winner-RoboCup-2022}. Their approach builds lines from points extracted in regions of the image that are classified as field lines. This classification relies on information from previous frames and characteristics of the current image.

Another recent research proposes using a CNN to perform object detection and semantic segmentation \cite{yoeo}. Their approach uses the same encoder for both tasks, reducing computational overheads for the robot's vision pipeline. It was trained with an extensive dataset, containing images taken in various real-world locations as well as a collection of simulated images \cite{TORSO2021}.

To optimize the detection of line segments on images, Suaréz et al. propose the Enhanced Line SEgment Drawing (ELSED) algorithm \cite{elsed}, which detects segments by connecting adjacent pixels with high gradients. Their approach joins the processes of edge drawing and segment detection in one single step, outperforming other efficient methods in terms of processing speed, such as the popular Line Segment Detector (LSD) \cite{LSD}. Even though it is less accurate than other Deep Learning-based competitors, its trade-off between accuracy and computational costs makes it ideal for resource-limited devices.


\section{Proposed Approach}\label{ch3}

\begin{figure}[b]
    \begin{center}
        \includegraphics[width=0.95\textwidth]{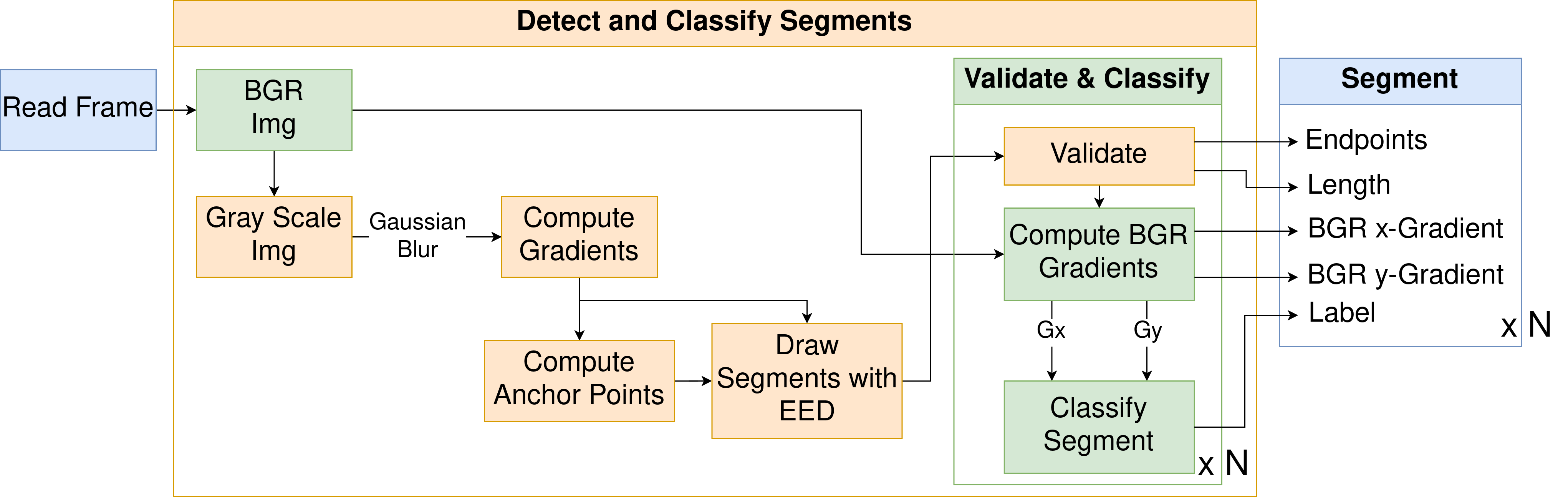}
    \end{center}
    \caption{Workflow of our proposed algorithm including high-level steps from ELSED.}
    \label{fig:algorithm}
\end{figure}

Figure \ref{fig:algorithm} illustrates our method's workflow. It begins by detecting line segments using the Enhanced Edge Drawing (EED) algorithm, which serves as the foundation for ELSED \cite{elsed}. In the validation step of ELSED, we analyze the average horizontal and vertical gradients of each RGB channel. By comparing the gradient vectors of detected segments with those expected from green-to-white transitions, which are characteristic of soccer field lines, we compute similarity metrics to identify the relevant segments.

The segment is classified as a field feature if the similarity metrics achieve predefined thresholds. These classification thresholds can be manually adjusted, but Section \ref{ch4} presents our method for automatic threshold tuning based on manually annotated images from the target environment using the PSO algorithm.

\subsection{The Enhanced Edge Drawing algorithm}


The first step of the EED algorithm computes a Gaussian smooth to suppress the image's noise. Next, it computes the image's horizontal and vertical gradients, $G_x$ and $G_y$, using the Sobel \cite{sobel} operator and calculates their $L1$ norms, $G = |G_x| + |G_y|$. It uses a threshold for discarding low gradient pixels and classify the remaining points into vertical or horizontal edges. The pixels with $G>0$ are scanned and compared against their surrounding gradients to find the local maxima called anchor points.

\begin{figure}[t]
    \begin{center}
        \includegraphics[width=0.9\textwidth]{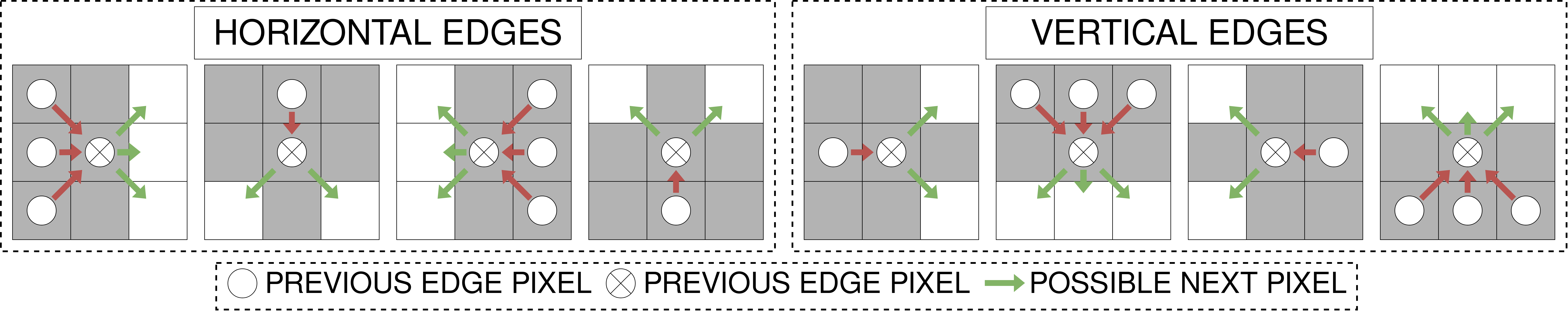}
    \end{center}
    \caption{Pixel selection for building line segments with EED \cite{elsed}.}
    \label{fig:eed}
\end{figure}

In the next step, EED performs edge drawing and line fitting simultaneously, reducing the number of checked pixels compared to other methods. They follow the assumption of Bresenham's line drawing \cite{bresenham}, proposing that diagonals' edge chains should form a line. As highlighted in Figure \ref{fig:eed}, the number of checked pixels can be 2 (diagonal cases) or 3 (non-diagonal cases), avoiding non-meaningful cases and accelerating the edge drawing algorithm since the typical edge drawing algorithm might check up to 6 pixels.

Whenever a change of orientation is detected, EED tries to continue in the same direction. The discontinuities are pushed into a stack, which is used later for discarding invalid segments. Discontinuities are skipped using the Bresenham's algorithm drawing in the current line segment direction and the next pixels are drawn following the edge direction from the first pixel after the discontinuity.

After dealing with discontinuities, to validate the line segments detected by the EED algorithm, ELSED computes the angular errors of each pixel's individual gradient orientation against the direction normal to the segment they belong to. In summary, a segment is validated if at least 50\% of its pixels have an angular error lower than a threshold.

\subsection{Gradient Extraction}
Since field lines are characterized by transitions from green to white, the pixels that compose the detected segments belonging to the field should have gradients that resemble this transition. The core idea of our proposed method is to compute the average gradient of the line segment and analyze its similarity with the gradient of a green-to-white transition in the RGB color space, which we will refer to as the $\vec{GW}$ vector.

The gradient operation using the Sobel filter is applied over a 3×3 window. Since we aim to compute the average gradient of the line segment, we can precompute the window averages for each pixel and then apply the gradient operation once at the end. This optimization reduces the computational cost, as the gradient computation is only performed once per segment.

We integrate gradient extraction into the validation step, which already iterates over the segments' pixels individually. While iterating through the pixels, we extract their 3×3 windows and compute the sum of their values. At the end of the segment, we divide the accumulated matrix by the number of pixels to obtain the average, and then compute its horizontal and vertical gradients using a Sobel operator. Resulting values from the convolutions are divided by four to scale them in the 255 range. This process is performed for all three channels of the RGB color space, ultimately yielding a three-dimensional vector for the horizontal gradient and another for the vertical gradient.

\begin{figure}[t]
    \begin{center}
        \includegraphics[width=0.95\textwidth]{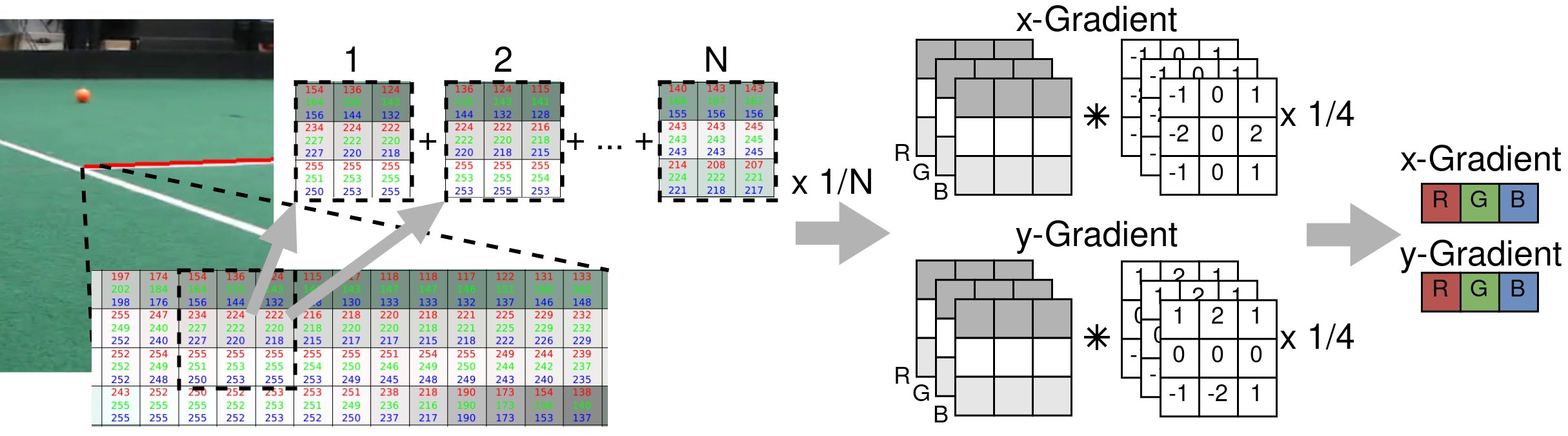}
    \end{center}
    \caption{Gradient extraction for classifying line segments.}
    \label{fig:grad}
\end{figure}

\subsection{Line Segment Classification}
\label{sub:classification}
To classify a segment, we compute the similarities between its RGB gradients and $\vec{GW}$. For that, we project the segment's gradient vector onto the $\vec{GW}$ vector. The length of this projection and the angle between the vectors serve as the primary metrics for classification. The projection length indicates the intensity of the color transition, while the angle measures whether the transition direction aligns correctly with the expected green-to-white shift.

In addition to the similarity metrics, we define a minimum length requirement for a segment to be considered a field feature. The computed similarity metrics are compared against predefined threshold values, and if all thresholds are met, the segment is classified as a field line.

After classification, we store the segment’s information in a structured format, which includes the segment's endpoints, length, horizontal and vertical gradients, and the resulting classification label, as show in Figure \ref{fig:algorithm}. These stored details are later used for segment visualization and for training the optimization of classification thresholds, a process that will be presented in Section \ref{ch4}.

Lastly, note that this approach can be employed for detecting any lines with known color transitions. For example, in the SSL, field boundaries are characterized by green to black transitions. Thus, we can compute lines' gradients similarity to a $\vec{GB}$ vector to check if they belong to a field boundary.

\section{Data Collection and Training} \label{ch4}
Our approach requires adjusting the classification thresholds for line segments. We developed a pipeline for annotating segment labels in images, enabling the use of optimization algorithms to find the best threshold values that maximize classification accuracy. We selected the Particle Swarm Optimization (PSO) algorithm for this task due to its versatility and fast convergence.

\subsection{Acquiring Data}

\begin{figure}[ht]
    \begin{center}
        \includegraphics[width=0.6\textwidth]{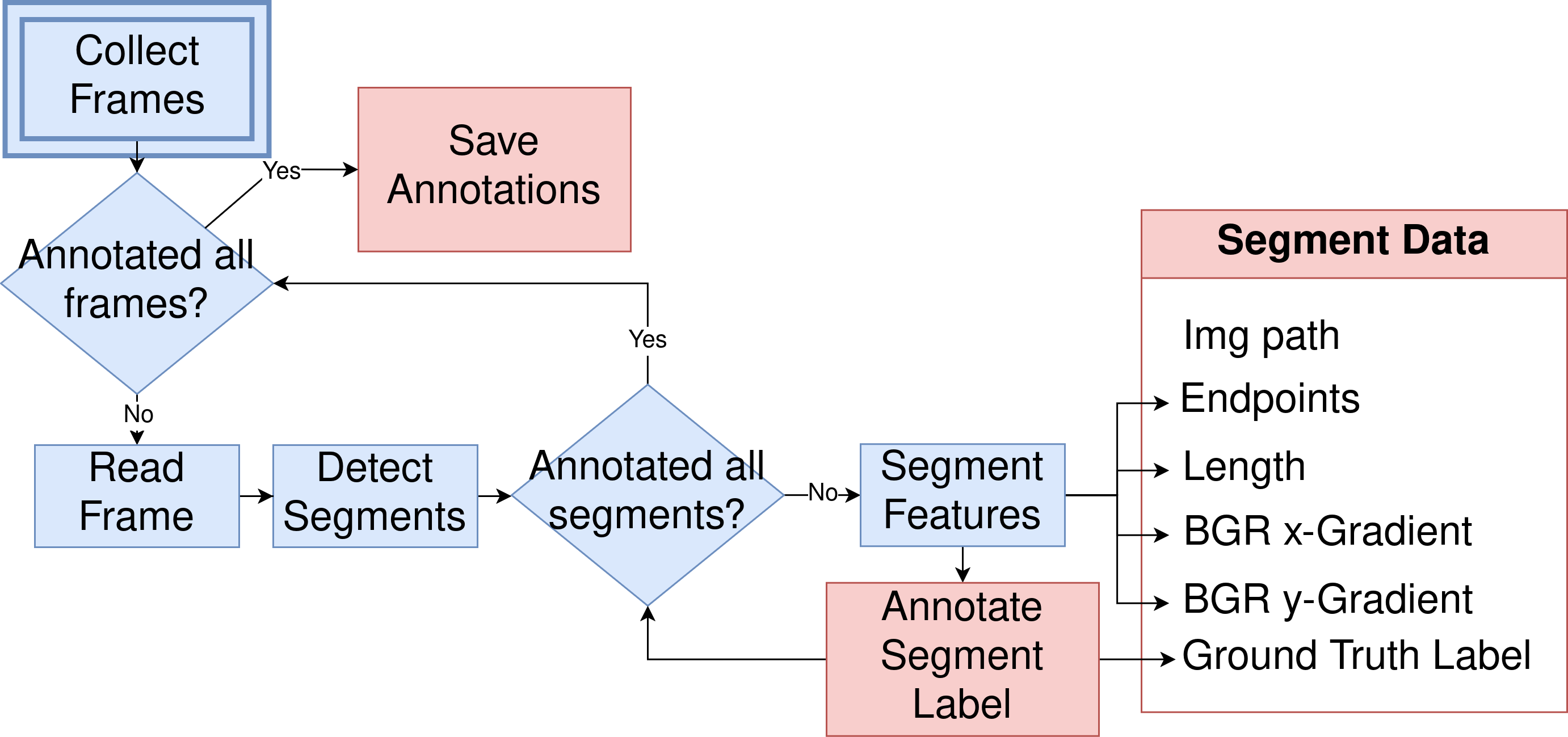}
    \end{center}
    \caption{Steps to annotate labels of line segments on images.}
    \label{fig:annotation}
\end{figure}

Once we have pictures acquired by the robot on the field, we follow the steps shown in Figure \ref{fig:annotation} to annotate segment labels on each image. First, we run our proposed method to extract gradients and visualize the detected line segments. Using these visualizations, we manually annotate each segment with a label indicating whether it belongs to a field line or not.

The annotations for each line segment are stored in a file containing the source image, segment endpoints, segment length, vertical and horizontal gradients, and the manually assigned label. This file can be accessed later to review and refine the annotations if needed.

\subsection{Training Thresholds}
\begin{figure}[ht]
    \begin{center}
        \includegraphics[width=1.0\textwidth]{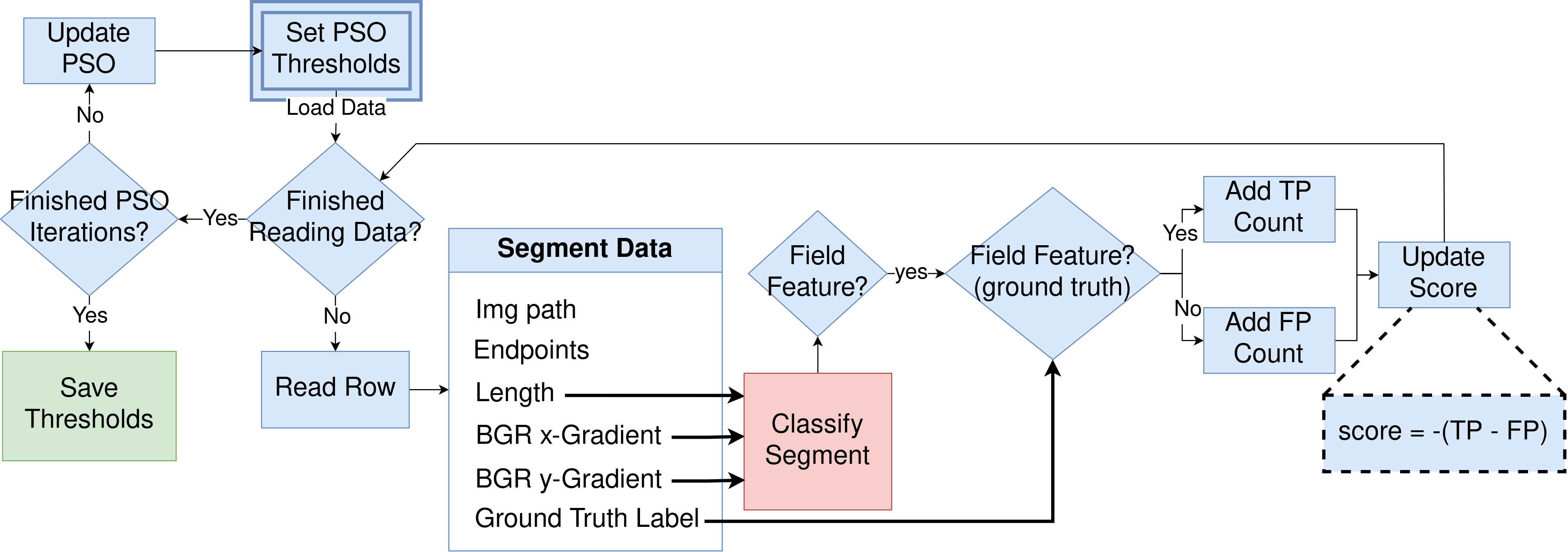}
    \end{center}
    \caption{Training pipeline to adjust classification thresholds.}
    \label{fig:training}
\end{figure}

PSO is used to optimize three thresholds: the maximum angle between the gradient and the $\vec{GW}$ vector, the minimum projection length of the gradient onto $\vec{GW}$, and the minimum segment length. Our training pipeline is shown in Figure \ref{fig:training}, which starts by defining a set of initial thresholds. New thresholds are generated and applied in each PSO iteration to classify the annotated segments.

We store the gradient values, segment length, and ground-truth label during the annotation phase. Using these annotated features, we classify each segment using the thresholds provided by PSO and compare the results with the ground-truth labels. The training process aims to maximize a score defined as the number of true positives (TP) minus the number of false positives (FP).

\section{Evaluation} \label{ch5}
Our evaluation experiments were conducted using the Jetson Orin Nano 8GB \cite{jetson-orin-nano}, a low-power platform designed for embedded systems development. Previous studies have shown that Jetson platforms are well-suited for implementing autonomous robots in the robot soccer environment \cite{towards-autonomous-ssl, humanoid:vision-pipeline}.


We compared our solution against a state-of-the-art semantic segmentation method, You Only Encode Once (YOEO) \cite{yoeo}. However, to enable a direct performance comparison, we added a line segment detection step after the segmentation generated by YOEO, using a probabilistic Hough Lines, as suggested by other researchers \cite{humanoid:vision-system}. Moreover, we've added a classification for field boundaries using the $\vec{GB}$ as reference, as suggested in Subsection \ref{sub:classification}. Ultimately, the methods were compared in terms of both accuracy and processing speed.

We also analyzed the behavior of our solution under different lighting conditions and variations in training set size. This analysis demonstrates that data collected under one lighting condition can be generalized to other conditions within the same environment. Moreover, we show that reducing the amount of training data does not significantly impact accuracy, proving that our approach requires only a small dataset to achieve effective results.

\subsection{Datasets}
We used five sets of images for our experiments. Two of these datasets are subsets of the TORSO-21 \cite{TORSO2021}, specifically containing images from the SPL and Humanoid Kid leagues. The remaining three datasets consist of images captured in different lighting conditions using an SSL-category robot in our laboratory. The characteristics of this robot have been detailed in previous studies \cite{towards-autonomous-ssl}.

Each dataset includes 30 evaluation images, while the remaining images were used for training and validating the thresholds of our method. To evaluate YOEO, we used its pre-trained weights in the TORSO-21 dataset. However, we did not fine-tune YOEO on our captured data, which allowed us to assess how well the model generalizes to new conditions not seen during training.

Table \ref{tab:datasets} presents the main characteristics of each dataset used in our experiments. From this point forward, we refer to the datasets using the labels from the table. For example, D2 refers to the dataset captured in our laboratory with both artificial and natural lighting.

\begin{table}[h]
\centering
\caption{Datasets characteristics.}
\label{tab:datasets}
\resizebox{1.0\textwidth}{!}{
\begin{tabular}{m{1.0cm}m{2.2cm}m{2.2cm}m{2.2cm}m{6.0cm}}
\hline
  \textbf{Label} &
  \textbf{Environment} &
  \textbf{Resolution} &
  \textbf{Nr of Images} &
  \multicolumn{1}{c}{\textbf{Description}} \\ \hline
  D1 &
  SSL &
  480x640 &
  140 &
  \begin{tabular}[c]{p{6.0cm}@{}l@{}}
  Lights turned off and windows open, only natural lighting 
  \end{tabular} 
  \\ \hline
  D2 &
  SSL &
  480x640 &
  190 &
  \begin{tabular}[c]{p{6.0cm}@{}l@{}}
  Lights turned on and windows open, combines natural and artificial lighting 
  \end{tabular} 
  \\ \hline
  D3 &
  SSL &
  480x640 &
  189 &
  \begin{tabular}[c]{p{6.0cm}@{}l@{}}
  Lights turned on and windows closed, only artificial lighting 
  \end{tabular} 
  \\ \hline
  D4 &
  Humanoid Kid &
  360x640 &
  74 &
  \begin{tabular}[c]{p{6.0cm}@{}l@{}}
  Images from multiple competition environments, no natural lighting 
  \end{tabular} 
  \\ \hline
  D5 &
  SPL &
  480x640 &
  86 &
  \begin{tabular}[c]{p{6.0cm}@{}l@{}}
  Images from a match in a closed environment, with people interfering and artificial lighting 
  \end{tabular} 
  \\ \hline
\end{tabular}
}
\end{table}

\subsection{Comparison with YOEO}

To run YOEO optimally on the Jetson Orin Nano, we converted the model into TensorRT format with INT8 precision. The model conversion and line detection implementation procedures are available in our repository \footnote{https://github.com/robocin/YOEO}.


To evaluate YOEO-based line detection accuracy, we applied the OpenCV implementation of the probabilistic Hough Transform to the segmented images produced by the model. Then, we ran this approach on the 30 evaluation images and manually annotated which lines belonged to the field and which did not, computing precision and recall metrics. Since the segmented images inherently filter non-field regions, the recall for YOEO-based detection is always 1.0. Our method was evaluated using the same procedure, processing the evaluation images and manually annotating the detected segments.


Table \ref{tab:accuracy_comparison} shows the precision and recall values for both methods across all datasets, along with the total number of line segments detected. First, we observe that our method achieved higher precision than the baseline in datasets D1, D2, and D3. This result can be attributed to the fact that the images from datasets D4 and D5 were used during the training of the YOEO model, allowing it to better handle those specific environments.

\begin{figure}[ht]
    \begin{center}
        \includegraphics[width=0.6\textwidth]{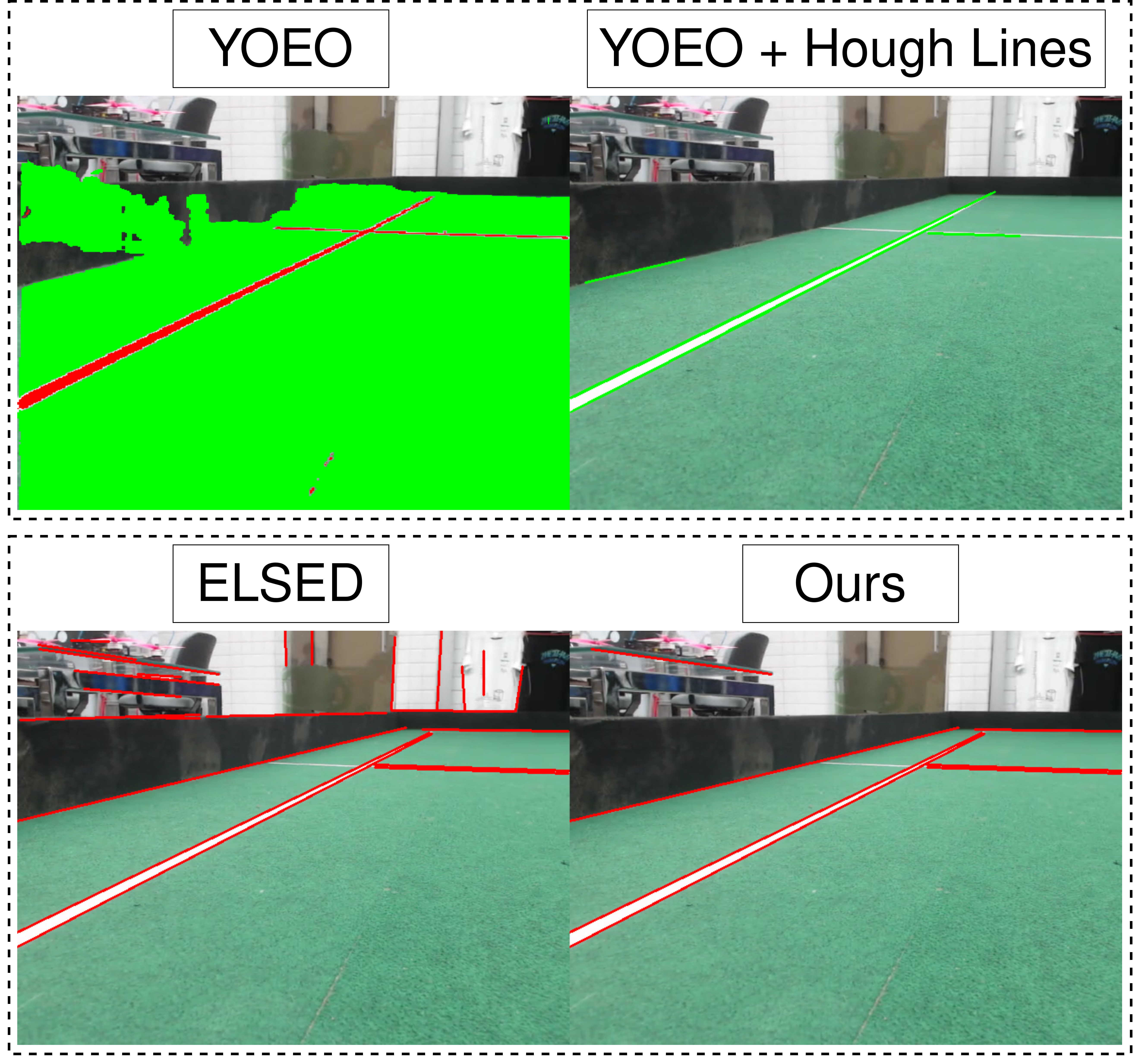}
    \end{center}
    \caption{Comparison illustrating how YOEO + Hough Lines and our method perform field lines detection.}
    \label{fig:comparison}
\end{figure}

We can see in Figure \ref{fig:comparison} that the segmentation obtained by YOEO incorrectly classifies some field pixels, which can lead to erroneous line detections. On the other hand, our method mistakenly classified a line that does not belong to the field due to its gradient having high similarity with that of the field lines. These are the main factors contributing to the errors of the evaluated methods.

Overall, both methods achieved high precision values, with the worst result observed for our method in the D5 dataset. This result can be explained by the significant presence of people in the scenes where the images were collected and the similarity in color between the Nao robot and the field lines. YOEO handles these adversities better by using a segmentation model over the entire image, analyzing the whole scene context to classify pixels rather than relying solely on information from line segments.

Table \ref{tab:accuracy_comparison} also presents mean and standard deviation processing time values, showing that our method was faster in all experiments. Additionally, since it runs on the CPU, it is more versatile for different platforms. In contrast, YOEO was evaluated running on a GPU in its most optimized form, which suggests that its performance might be slower on platforms without a GPU. However, YOEO performs both segmentation and object detection within the same model.

\begin{table}[t]
\centering
\caption{Accuracy and processing time results.}
\label{tab:accuracy_comparison}
\resizebox{1\textwidth}{!}{%
\begin{tabular}{ccccccccc}
\hline
\multirow{2}{*}{\textbf{Label}} & \multicolumn{4}{c}{\textbf{YOEO + Hough Lines}} & \multicolumn{4}{c}{\textbf{Ours}} \\ \cline{2-9} 
& precision & recall & nr of lines & processing time & precision & recall & nr of lines & processing time \\ \hline
D1   & 0.948          & 1.0 & 191 & $22.8 \pm 1ms$ & \textbf{1.0}   & 0.737 & 175 & $\boldsymbol{17.6 \pm 4ms}$ \\ \hline
D2  & 0.965          & 1.0 & 228 & $23.6 \pm 2ms$ & \textbf{0.994} & 0.903 & 196 & $\boldsymbol{16.8 \pm 2ms}$ \\ \hline
D3 & 0.939          & 1.0 & 230 & $25.1 \pm 2ms$ & \textbf{0.987} & 0.847 & 185 & $\boldsymbol{18.3 \pm 3ms}$ \\ \hline
D4  & \textbf{1.0}   & 1.0 & 77  & $25.2 \pm 2ms$ & \textbf{1.0}   & 0.453 & 243 & $\boldsymbol{13.7 \pm 6ms}$ \\ \hline
D5   & \textbf{0.942} & 1.0 & 530 & $24.0 \pm 2ms$ & 0.875          & 0.678 & 453 & $\boldsymbol{20.6 \pm 5ms}$ \\ \hline
\end{tabular}%
}
\end{table}

\subsection{Tests with Different Illuminations and Training Set Sizes}

\begin{figure}[ht]
    \begin{center}
        \includegraphics[width=1\textwidth]{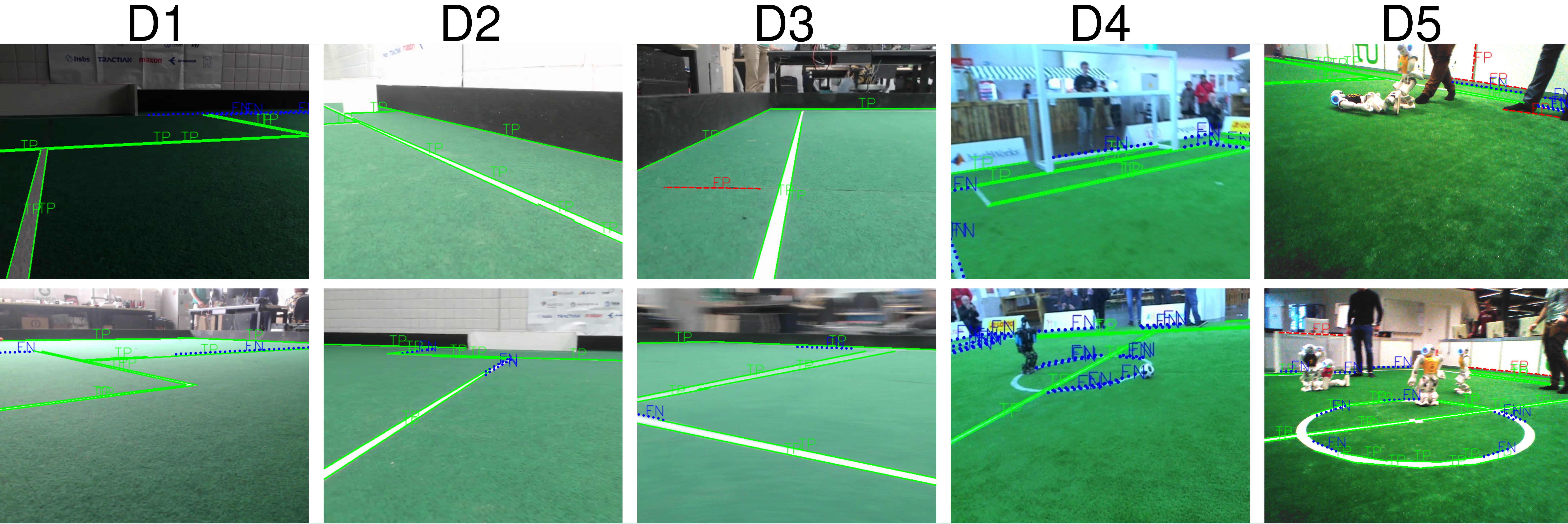}
    \end{center}
    \caption{Examples of detections from our method on each dataset.}
    \label{fig:examples}
\end{figure}

We conducted experiments using different training set sizes to demonstrate that our method does not require large amounts of data to achieve accurate results. Results are shown in Table \ref{tab:set-sizes}, where the “Set Size” column indicates the percentage of the dataset used for training in each evaluation. 

As examples, Figure \ref{fig:examples} illustrates detections performed by our method in two images of each dataset, highlighting true positives (green continuous lines), false positives (red dashed lines), and false negatives (blue dotted lines).

The results show that even with reduced training data, the algorithm maintains a high level of precision. In some cases, the method even performs better with less data, a behavior that can be attributed to the randomness inherent in the PSO-based training process. This low data requirement and ease of labeling in our method presents a notable advantage over CNN-based semantic segmentation approaches, which typically demand extensive annotated datasets.


\begin{table}[h]
\centering
\caption{Precision and recall with different training set sizes.}
\label{tab:set-sizes}
\resizebox{0.9\textwidth}{!}{%
\begin{tabular}{ccccccccccc}
\hline
\multirow{2}{*}{\textbf{Set Size}} & \multicolumn{2}{c}{\textbf{D1}} & \multicolumn{2}{c}{\textbf{D2}} & \multicolumn{2}{c}{\textbf{D3}} & \multicolumn{2}{c}{\textbf{D4}} & \multicolumn{2}{c}{\textbf{D5}} \\ \cline{2-11} 
 & precision & recall & precision & recall & precision & recall & precision & recall & precision & recall \\ \hline
10\% & 0.88 & 0.84 & 1.0  & 0.69 & 0.98 & 0.84 & 0.78 & 0.44 & 0.79 & 0.81 \\ \hline
20\% & 0.95 & 0.67 & 1.0  & 0.79 & 0.95 & 0.77 & 0.94 & 0.72 & 0.79 & 0.81 \\ \hline
30\% & 0.98 & 0.90 & 0.99 & 0.90 & 0.98 & 0.75 & 0.96 & 0.59 & 0.88 & 0.63 \\ \hline
40\% & 0.96 & 0.80 & 1.0  & 0.81 & 0.98 & 0.75 & 0.96 & 0.56 & 0.79 & 0.81 \\ \hline
50\% & 0.94 & 0.78 & 0.99 & 0.81 & 0.97 & 0.83 & 0.96 & 0.65 & 0.9 & 0.74 \\ \hline
60\% & 0.99 & 0.85 & 1.0  & 0.77 & 0.96 & 0.91 & 0.96 & 0.55 & 0.82 & 0.79 \\ \hline
70\% & 0.99 & 0.79 & 1.0  & 0.74 & 0.97 & 0.87 & 0.93 & 0.67 & 0.88 & 0.74 \\ \hline
\end{tabular}%
}
\end{table}

We also performed experiments to assess the robustness of our approach under different lighting conditions. For this purpose, we conducted cross-evaluations using thresholds trained on datasets D1, D2, and D3. We measured the precision obtained in each case, applying the thresholds to images acquired under different lighting setups.

\begin{figure}[ht]
    \begin{center}
        \includegraphics[width=0.55\textwidth]{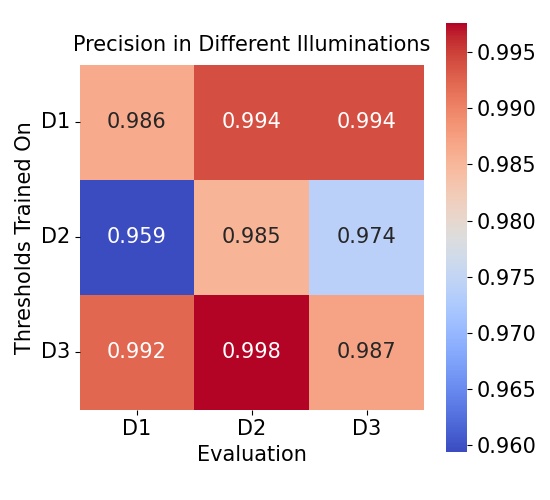}
    \end{center}
    \caption{Precision from evaluations in different illuminations.}
    \label{fig:illuminations}
\end{figure}

The results, shown in Figure~\ref{fig:illuminations}, indicate that changes between natural and artificial lighting did not significantly impact the method’s accuracy. In all evaluations, the method achieved precision levels above 96\%, with very similar values across the board. This result highlights the robustness of the proposed approach in handling illumination variations without the need for re-training.

\section{Conclusion} \label{ch6}
This paper proposes a real-time field line detection approach designed for low-computation platforms in robot soccer environments. Our method detects line segments using the ELSED algorithm and classifies them based on RGB color transitions observed along the pixels that constitute each segment. By computing the similarity between the segment’s gradient and the expected color transitions of field features, we identify the segments that belong to the field.

Our approach also relies on threshold selection for classification. To this end, we introduced a complete pipeline for label annotation and threshold training using Particle Swarm Optimization (PSO). This feature makes the process of annotation and threshold tuning fast and straightforward, facilitating easy adaptation to different competition environments.

Experimental results demonstrate that our method achieves accuracy comparable to the state-of-the-art YOEO-based field line detector. Furthermore, it outperforms YOEO in terms of processing speed—even when YOEO runs in its most optimized GPU configuration using INT8 precision.

While YOEO provides robust detection through a CNN-based architecture, it may require retraining when used in environments that differ significantly from those in its training dataset. Such retraining demands a large set of labeled data, which may be impractical during competitions. In contrast, our method offers a lightweight and adaptable alternative that can be quickly tuned to new conditions.

In future work, we plan to explore unsupervised clustering of segment gradients during training, aiming to eliminate the need for manual annotations and enable a fully unsupervised training process.


%
%
%
\bibliographystyle{splncs04}

\bibliography{main}

\end{document}